\pdfoutput=1

\documentclass[11pt]{article}

\usepackage[]{ACL2023}

\usepackage{times}
\usepackage{latexsym}

\usepackage[T1]{fontenc}

\usepackage[utf8]{inputenc}

\usepackage{microtype}

\usepackage{inconsolata}

\usepackage{CJKutf8}
\usepackage{multirow}
\usepackage{amssymb}
\usepackage{CJKutf8}
\usepackage{graphicx}
\usepackage{hyperref}

\title{TCBERT: A Technical Report for Chinese Topic Classification BERT}

\author{Ting Han$^1$, Kunhao Pan$^1$, Xinyu Chen$^{2}\footnotemark[1]$, Dingjie Song$^{3}\footnotemark[1]$, Yuchen Fan$^1$, \\ {\bf Xinyu Gao$^1$, Ruyi Gan$^1$ \and Jiaxing Zhang$^1$}\\
  $^1$International Digital Economy Academy (IDEA), China \\
  $^2$Beijing University of Posts and Telecommunications, China \\
  $^3$Nanjing University, China \\
  \texttt{\{hanting,pankunhao,fanyuchen,gaoxinyu,ganruyi,zhangjiaxing\}@idea.edu.cn} \\
          \texttt{chenxiny@bupt.edu.cn}, \texttt{songdj@smail.nju.edu.cn}}

\begin{document}
\maketitle

\renewcommand{\thefootnote}{\fnsymbol{footnote}}
\footnotetext[1]{Contributed to the work as interns at IDEA.}

\renewcommand{\thefootnote}{\arabic{footnote}}

\begin{abstract}
Bidirectional Encoder Representations from Transformers or BERT~\cite{devlin-etal-2019-bert} has been one of the base models for various NLP tasks due to its remarkable performance. Variants customized for different languages and tasks are proposed to further improve the performance. In this work, we investigate supervised continued pre-training~\cite{gururangan-etal-2020-dont} on BERT for Chinese topic classification task. Specifically, we incorporate prompt-based learning and contrastive learning into the pre-training. To adapt to the task of Chinese topic classification, we collect around 2.1M Chinese data spanning various topics. The pre-trained Chinese Topic Classification BERTs (TCBERTs) with different parameter sizes are open-sourced at \url{https://huggingface.co/IDEA-CCNL}.
\end{abstract}

\begin{CJK}{UTF8}{gbsn}
\section{Introduction}
Deep bidirectional neural networks based on Transformers~\cite{vaswani2017attention} have been one of the prevalent structures to encode natural language. BERT~\cite{devlin-etal-2019-bert}, as the most representative bidirectional encoder, learns language representations through pre-training on massive text data using the masked language modeling (MLM) objective, a variant of Cloze task~\cite{taylor1953cloze}. The pre-trained representations achieve remarkable performance across a wide range of NLP tasks through fine-tuning on small specific datasets. The pre-train and fine-tune paradigm of BERT inspires multiple follow-up research work on different tasks~\cite{joshi-etal-2020-spanbert, yin-etal-2020-tabert, wu-etal-2020-tod, ji-etal-2021-spellbert}, and also different languages~\cite{martin-etal-2020-camembert, sun-etal-2021-chinesebert, abdul-mageed-etal-2021-arbert}. 

Recently,~\citet{gururangan-etal-2020-dont} demonstrate that continued pre-training on specific domains or tasks before fine-tuning consistently advances the performance of related tasks. Similarly, in this work, we investigate the continued pre-training for the task of Chinese topic classification on BERT. In addition to the MLM objective used in the continued pre-training, we incorporate prompt-based learning to leverage labeled topic information, and contrastive learning to improve sentence representations into the supervised continued pre-training. Specifically, the MLM is only used to predict label information masked in the prompt template while the other input tokens remain intact. To construct positive pairs, a text sentence is paired with its prompt-appended version for contrastive learning. To adapt to Chinese topic classification tasks, we collect 2.1M annotated Chinese data across diverse topics. We evaluate the pre-trained Chinese topic classification BERT (TCBERT) on three few-shot datasets, TNEWS, CSLDCP and IFLYTEK of FewCLUE~\cite{xu2021fewclue}, and the experimental results are presented for reference.

\section{Related Work}

\begin{figure*}[t]
  \centering
  \includegraphics[width=\textwidth]{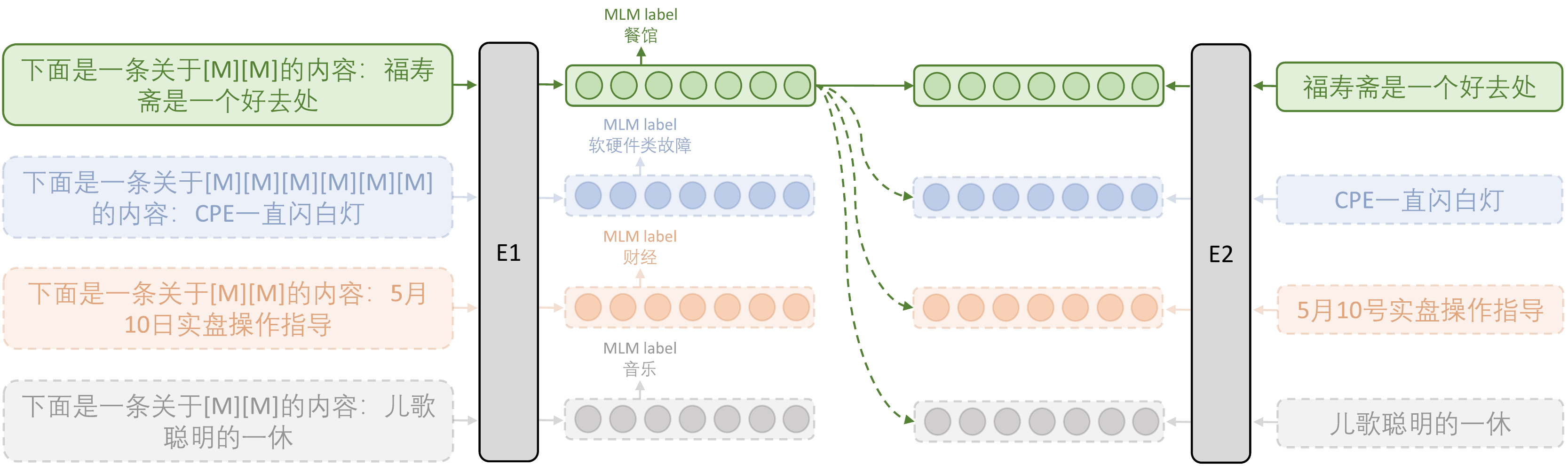}
  \caption{(a) The left part presents the prompt-based mask language modeling only using encoder E1. (b) The whole figure presents the prompt-based contrastive learning using both encoders E1 and E2. The solid line denotes positive contrastive pairs while the dash lines indicate negative ones.}
  \label{fig:main}
\end{figure*}

\subsection{Supervised Pre-training}

Most pre-training approaches~\cite{raffel2020exploring, lewis-etal-2020-bart, liu2019roberta, devlin-etal-2019-bert} adopt unsupervised learning on large-scale general corpus to bypass data annotation procedures which are costly in supervised learning. With more large-scale annotated datasets have become accessible, a few works~\cite{worsham2020multi, aghajanyan-etal-2021-muppet, dosovitskiy2020image} show that supervised pre-training can also achieve competitive or even better performance compared to unsupervised pre-training. In this work, we conduct continued supervised pre-training based on models pre-trained by unsupervised learning for Chinese topic classification using around 2.1M annotated Chinese topic data.

\subsection{Contrastive Learning}
Contrastive learning, first proposed by~\citet{hadsell2006dimensionality}, has been successfully extended in computer vision~\cite{pmlr-v119-chen20j, he2020momentum} and natural language processing~\cite{gao-etal-2021-simcse, meng2021coco} for representation learning. The core of contrastive learning to learn representations is to form contrastive pairs. In NLP, multiple operations ~\cite{gao-etal-2021-simcse, zhang-etal-2021-shot, fang2020cert, wu2020clear, yan-etal-2021-consert, wang-etal-2021-cline} are proposed to construct the contrastive pairs. In applying contrastive learning to the supervised continued pre-training, we are mainly inspired by SimCSE~\cite{gao-etal-2021-simcse} and MOCO~\cite{he2020momentum}.

\subsection{Prompt-based Learning}

Natural language descriptions or task descriptions are applied to solve either few-shot~\cite{schick-schutze-2021-exploiting, brown2020language} or zero-shot~\cite{radford2019language, wei2021finetuned} NLP tasks. Since then, a new paradigm~\cite{liu2021pre} termed \textit{prompt-based learning} has been a trendy research topic in the NLP community. On prompt design, continuous prompts~\cite{li-liang-2021-prefix, vu-etal-2022-spot, gu-etal-2022-ppt, liu2021gpt} are easy to be combined with gradient descent but uninterruptible to humans~\cite{khashabi-etal-2022-prompt} while discrete prompts~\cite{tam-etal-2021-improving, petroni-etal-2019-language, schick-schutze-2021-exploiting, jiang-etal-2020-know, gao-etal-2021-making, liu-etal-2022-makes} are human-readable but introduce an extra step to either manually create or automatically generate the prompts. In this work, we mainly follow the format of manually created prompts in PET~\cite{schick-schutze-2021-exploiting} for supervised continued pre-training.

\section{Topic Classification BERT}
\subsection{Model Structure}
Topic classification BERT or TCBERT taking a similar structure of BERT is shown in Figure~\ref{fig:main}. We pre-train TCBERT with three different sizes: TCBERT-base of 110M parameters, TCBERT-large of 330M parameters and TCBERT-1.3B~\cite{shoeybi2019megatron} of 1.3B parameters. The pre-training objectives which are based on prompts are detailed in the following sections.
\subsection{Prompt-based Mask Language Modeling}
As shown in Figure~\ref{fig:main}, each input sentence $sent_{i}$ consists of a prompt template and a text sentence $u_{i}$. Prompt-based mask language modeling is to output the [MASK] tokens in the input sentence to the label of the text sentence. For instance, TCBERT will output "餐馆" for the [MASK] tokens in the input sentence (top one in green in Figure~\ref{fig:main}) as "餐馆" is the label of the text sentence. The MLM loss is computed as follows:

\begin{equation}
    \mathcal{L}_{MLM} = -\frac{1}{N} \sum_{m=1}^{M} \log P(x_{m}),
\end{equation}
where M is the total number of masked input tokens and $P(x_{m})$ is the predicted probability of the masked token $x_{m}$ over the vocabulary.

\subsection{Prompt-based Contrastive Learning}
Inspired by SimCSE~\cite{gao-etal-2021-simcse}, we introduce a contrastive training objective in addition to the $\mathcal{L}_{MLM}$ loss:
\begin{equation}
    \mathcal{L}_{CL}=-\sum_{i=1}^{N} \log \frac{\exp( \mathrm{sim}(\textbf{h}_{i}, \hat{\textbf{h}}_{i})/ \tau)}{\sum_{j=1}^{N} \exp(\mathrm{sim}(\textbf{h}_{i}, \hat{\textbf{h}}_{j})/ \tau)},
\end{equation}
where $\tau$ is a temperature hyperparameter and $N$ is the number of input sentences in a batch. $\mathrm{sim}(\textbf{h}_{i}, \hat{\textbf{h}}_{i})$ is the cosine similarity score of the representations $\textbf{h}_{i}$ and $\hat{\textbf{h}}_{i}$. As shown in Figure~\ref{fig:main}, positive pairs for contrastive learning are the text sentence $u_{i}$ and its prompt version $sent_{i}$. In this case, $\textbf{h}_{i}$ and $\hat{\textbf{h}}_{i}$ denotes the representations of $sent_{i}$ and $u_{i}$, respectively. To encode the sentences, we try both single-encoder and dual-encoder approaches for pre-training. Similar to SimCSE, we first try to encode the pairs using a single encoder. For the dual-encoder, we try MOCO~\cite{he2020momentum} and use the momentum encoder to encode the text sentences $u_{i}$. As in Figure~\ref{fig:main}, denoting E2 as the momentum encoder and E1 as the normal encoder, the parameters of E2 are updated as follows:
\begin{equation}
    \theta_{E2} = \lambda_{m} \theta_{E2} + (1-\lambda_{m}) \theta_{E1}, 
\end{equation}
where $\theta_{E2}$ and $\theta_{E1}$ are the parameters of E2 and E1, and $\lambda_{m} \in [0, 1)$ is a momentum coefficient. Note that only the parameters $\theta_{E1}$ are updated by back-propagation during pre-training. The two losses are equally weighted for pre-training: 
\begin{equation}
\mathcal{L}_{PCL} = \mathcal{L}_{MLM} + \mathcal{L}_{CL}.
\end{equation}
\section{Experimental Setups}
\subsection{Data}
To pre-train TCBERT, we collect around 2.1M Chinese data annotated with various topics including travel, movie, finance, etc. Three evaluation datasets, TNEWS, CSLDCP and IFLYTEK, are adopted from FewCLUE~\cite{xu2021fewclue} to demonstrate the performance of TCBERT.
\begin{table*}
\centering
\begin{tabular}{l|c|c|c|c|c|c}
\hline
\multirow{2}{*}{Model} & \multicolumn{3}{c|}{Prompt-1} & \multicolumn{3}{c}{Prompt-2} \\ \cline{2-7} 
                       & TNEWS   & CLSDCP   & IFLYTEK  & TNEWS   & CSLDCP  & IFLYTEK  \\ \hline
Macbert-base            & 55.02 & 57.37 & 51.34 & 54.78 & 58.38 & 50.83    \\
Macbert-large          & 55.77 & 58.99 & 50.31 & 56.77 & 60.22 & 51.63    \\
Erlangshen-1.3B          & 57.36 & 62.35 & 53.23 & 57.81 & 62.80 & 52.77   \\ \hline
TCBERT-base$^\diamondsuit$            & 55.57   & 58.60    & 49.63    & 54.58   & 59.16   & 49.80    \\
TCBERT-large$^\diamondsuit$           & 56.17   & 60.06    & 51.34    & 56.22   & 61.23   & 50.77    \\
TCBERT-1.3B$^\diamondsuit$            & 57.41   & 65.10    & 53.75    & 57.41   & 64.82   & 53.34    \\ \hline
TCBERT-base$^\clubsuit$            & 55.47   & 59.61    & 50.20    & 54.33   & 59.72   & 50.77    \\
TCBERT-large$^\clubsuit$           & 55.27   & 61.34    & 51.63    & 55.62   & 60.90   & 50.60    \\
TCBERT-1.3B$^\clubsuit$            & 56.92   & 63.31    & 52.43    & 56.92   & 63.14   & 52.83    \\ \hline
TCBERT-base$^\spadesuit$            & 54.68   & 59.78    & 49.40    & 54.68   & 59.78   & 49.40    \\
TCBERT-large$^\spadesuit$           & 55.32   & 62.07    & 51.11    & 55.32   & 62.07   & 51.11    \\
TCBERT-1.3B$^\spadesuit$            & 57.46   & 65.04    & 53.06    & 56.87   & 65.83   & 52.94    \\ \hline
\end{tabular}
\caption{Testing results of fine-tuning TCBERTs. $\diamondsuit$, $\clubsuit$ and $\spadesuit$ denote the TCBERTs are pre-trained with prompt-based MLM, prompt-based contrastive learning and prompt-based MOCO objectives, respectively.}
\label{tab:main-finetune}
\end{table*}

\subsection{Data Pre-processing}
\begin{figure}[t]
  \centering
  \includegraphics[width=1.0\columnwidth]{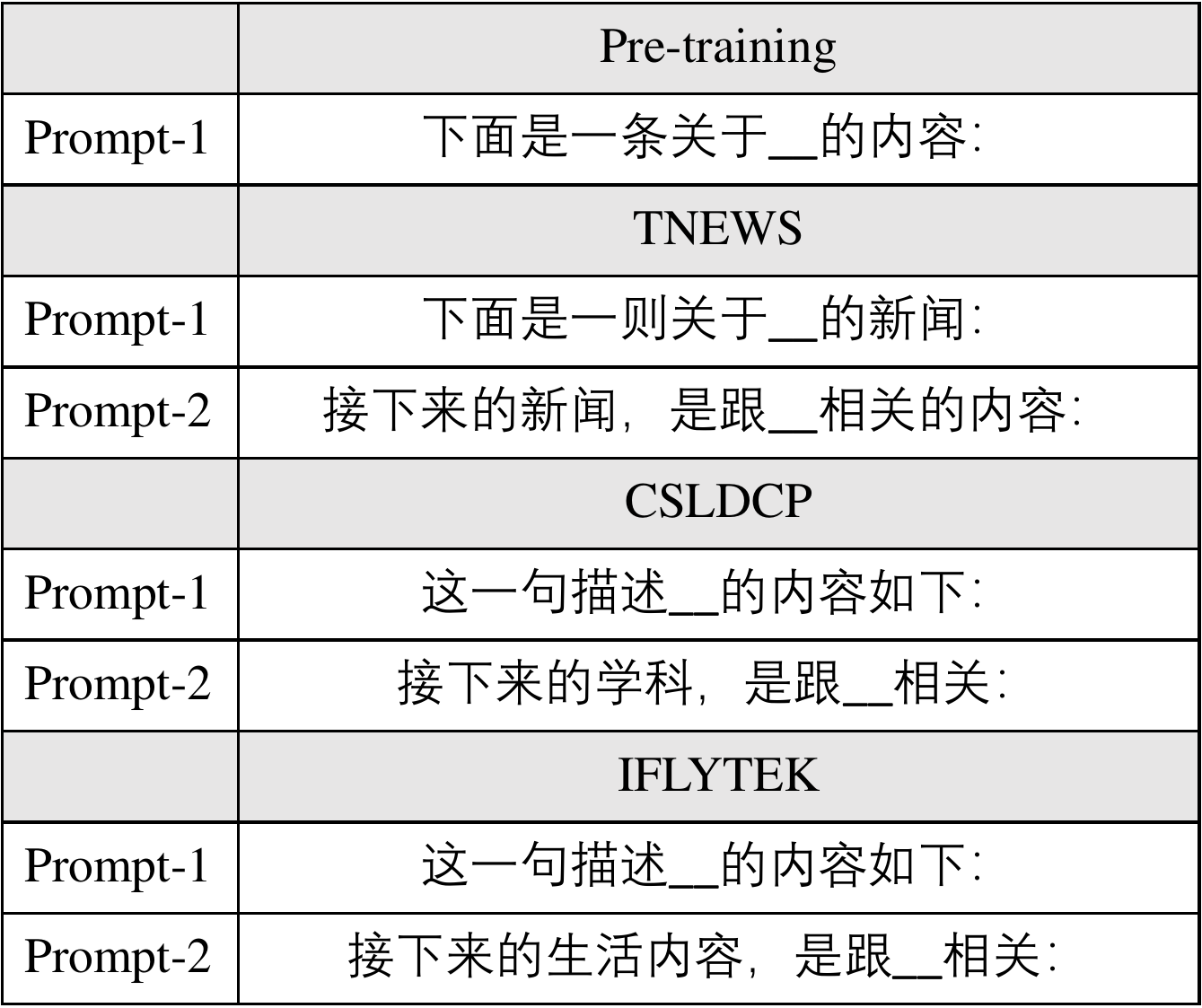}
  \caption{Prompt templates for pre-training and fine-tuning.}
  \label{fig:prompt}
\end{figure}

Prompt templates for pre-training and fine-tuning TCBERT are listed in Figure~\ref{fig:prompt}. For pre-training, we use one prompt template for all data. If one data sample is annotated with multiple topic labels, we process the data into multiple input sentences but with different topic labels for MLM predictions. For fine-tuning, we slightly modify the pre-training prompt template according to the topic of each evaluation datasets. Inspired by~\citet{webson-pavlick-2022-prompt}, we devise \textit{prompt-2} that is different from \textit{prompt-1} and includes more punctuation marks.

\subsection{Prompt-based Fine-tuning}
Similar to the pre-training stage, the prompt-based mask language modeling, $\mathcal{L}_{MLM}$, is also adopted for fine-tuning. In addition, TCBERT is fine-tuned under supervised signals for classification as follows:  
\begin{equation}
    \mathcal{L}_{TC} = -\frac{1}{N} \sum_{j=1}^{T} \sum_{i=1}^{N} \log P(T_{j}|sent_{i}),
\end{equation}
where $P(T_{j}|sent_{i})$ denotes predicted probability of the input sentence $sent_{i}$ belonging to the topic class $T_{j}$. We jointly train the two losses:
\begin{equation}
    \mathcal{L}_{PTC} = \mathcal{L}_{MLM} + \mathcal{L}_{TC},
\end{equation}
and the two losses are equally weighted during fine-tuning.

\subsection{Prompt-based Sentence Similarity}
Instead of fine-tuning the parameters of TCBERT for topic classification, we directly classify each test sample based on sentence similarity. The similarity-based method is different from zero-shot learning since the inference is performed using all training samples, and the parameters are not required to be updated by training samples, which is the typical manner of few-shot learning. The classification metric is as follows:
\begin{equation}
    score_{m,k} = \cos(\mathrm{R}(sent_{m}), \mathrm{R}(sent_k)),
\end{equation}
where $\mathrm{R}(\cdot)$ is the pooling method to extract sentence representation. Specifically, we average the last hidden layers of TCBERT as the sentence representation. $\cos$ denotes the cosine similarity of two sentence representations, and the $score_{m,k}$ is the similarity score between the testing sentence $sent_{m}$ and the training sentence $sent_{k}$.

The topic label assigned to the testing sentence $sent_{m}$ is according to the following rules:
\begin{equation}
    label_{sent_m} = label_{sent_{p}},
\end{equation}
\begin{equation}
    p = \arg\max_{k \in K}(score_{m,k}),
\end{equation}
where $sent_{p}$ is a sample from a training set with $K$ total number of training samples. Note that the sentence representation is computed including prompts. For samples in the training set, the topic labels are included in the prompts while for samples in the testing set, we use [MASK] tokens to represent topic labels in the prompts.

\subsection{Training Details}

TCBERTs with different sizes of parameters are initialized from MacBERT-base~\cite{cui-etal-2020-revisiting}, MacBERT-large and Erlangshen-MegatronBert-1.3B~\cite{wang2022fengshenbang}, respectively. The training parameters are optimized by AdamW~\cite{loshchilov2017decoupled} with a learning rate of 1e-5. We use a warmup rate of 0.001 and a weight decay of 0.1. TCBERTs are pre-trained 4 epochs with a batch size of 128 for TCBERT-base, and 32 for both TCBERT-large and TCBERT-1.3B. A maximum sequence length of 128 is applied to all pre-trainings. We fine-tune TCBERTs 50 epochs with the same learning rate. The batch size for fine-tuning is set to 4 for TNEWS and 2 for both CLSDCP and IFLYTEK with 512 as the maximum number of input tokens. All pre-training experiments are conducted on a single A100 GPU with 80GB memory, and the fine-tuning experiments are run on a single A100 GPU with 40G memory.

\subsection{TCBERTs at Hugging Face}
\begin{table}[t]
\centering
\resizebox{\columnwidth}{!}{
\begin{tabular}{c}
\hline
Hugging Face Model Cards                                    \\ \hline
\href{https://huggingface.co/IDEA-CCNL/Erlangshen-TCBert-110M-Classification-Chinese}{IDEA-CCNL/Erlangshen-TCBert-110M-Classification-Chinese}     \\
\href{https://huggingface.co/IDEA-CCNL/Erlangshen-TCBert-330M-Classification-Chinese}{IDEA-CCNL/Erlangshen-TCBert-330M-Classification-Chinese}     \\
\href{https://huggingface.co/IDEA-CCNL/Erlangshen-TCBert-1.3B-Classification-Chinese}{IDEA-CCNL/Erlangshen-TCBert-1.3B-Classification-Chinese}    \\ \hline
\href{https://huggingface.co/IDEA-CCNL/Erlangshen-TCBert-110M-Sentence-Embedding-Chinese}{IDEA-CCNL/Erlangshen-TCBert-110M-Sentence-Embedding-Chinese} \\
\href{https://huggingface.co/IDEA-CCNL/Erlangshen-TCBert-330M-Sentence-Embedding-Chinese}{IDEA-CCNL/Erlangshen-TCBert-330M-Sentence-Embedding-Chinese} \\
\href{https://huggingface.co/IDEA-CCNL/Erlangshen-TCBert-1.3B-Sentence-Embedding-Chinese}{IDEA-CCNL/Erlangshen-TCBert-1.3B-Sentence-Embedding-Chinese} \\ \hline
\end{tabular}
}
\caption{Hugging Face model cards of six TCBERTs.}
\label{tab:tcbert}
\end{table}
We open-source six pre-trained TCBERTs with different parameter sizes at Hugging Face. The model cards are listed in Table~\ref{tab:tcbert}. "Classification" denotes the pre-training objective is prompt-based MLM and "Sentence-Embedding" is prompt-based MOCO.
\begin{table*}[t]
\centering
\begin{tabular}{l|cccccc}
\hline
\multirow{3}{*}{Model} & \multicolumn{6}{c}{Prompt-1}                                                                                                                                                  \\ \cline{2-7} 
                       & \multicolumn{2}{c|}{TNEWS}                                     & \multicolumn{2}{c|}{CSLDCP}                                     & \multicolumn{2}{c}{IFLYTEK}                \\ \cline{2-7} 
                       & \multicolumn{1}{c|}{referece} & \multicolumn{1}{c|}{whitening} & \multicolumn{1}{c|}{reference} & \multicolumn{1}{c|}{whitening} & \multicolumn{1}{c|}{reference} & whitening \\ \hline
Macbert-base           & \multicolumn{1}{c|}{43.53}    & \multicolumn{1}{c|}{47.16}     & \multicolumn{1}{c|}{33.50}     & \multicolumn{1}{c|}{36.53}     & \multicolumn{1}{c|}{28.99}     & 33.85     \\
Macbert-large          & \multicolumn{1}{c|}{46.17}    & \multicolumn{1}{c|}{49.35}     & \multicolumn{1}{c|}{37.65}     & \multicolumn{1}{c|}{39.38}     & \multicolumn{1}{c|}{32.36}     & 35.33    \\
Erlangshen-1.3B           & \multicolumn{1}{c|}{45.72}    & \multicolumn{1}{c|}{49.60}     & \multicolumn{1}{c|}{40.56}     & \multicolumn{1}{c|}{44.26}     & \multicolumn{1}{c|}{29.33}     & 36.48     \\ \hline
TCBERT-base$^\diamondsuit$            & \multicolumn{1}{c|}{48.61}    & \multicolumn{1}{c|}{51.99}     & \multicolumn{1}{c|}{43.31}     & \multicolumn{1}{c|}{45.15}     & \multicolumn{1}{c|}{33.45}     & 37.28     \\
TCBERT-large$^\diamondsuit$           & \multicolumn{1}{c|}{50.50}    & \multicolumn{1}{c|}{52.79}     & \multicolumn{1}{c|}{52.89}     & \multicolumn{1}{c|}{53.89}     & \multicolumn{1}{c|}{34.93}     & 38.31     \\
TCBERT-1.3B$^\diamondsuit$            & \multicolumn{1}{c|}{50.80}    & \multicolumn{1}{c|}{51.59}     & \multicolumn{1}{c|}{51.93}     & \multicolumn{1}{c|}{54.12}     & \multicolumn{1}{c|}{33.96}     & 38.08     \\ \hline
TCBERT-base$^\clubsuit$            & \multicolumn{1}{c|}{48.16}    & \multicolumn{1}{c|}{51.54}     & \multicolumn{1}{c|}{46.55}     & \multicolumn{1}{c|}{49.30}     & \multicolumn{1}{c|}{35.33}     & 37.74     \\
TCBERT-large$^\clubsuit$           & \multicolumn{1}{c|}{48.51}    & \multicolumn{1}{c|}{50.20}     & \multicolumn{1}{c|}{50.31}     & \multicolumn{1}{c|}{50.08}     & \multicolumn{1}{c|}{36.82}     & 38.08     \\
TCBERT-1.3B$^\clubsuit$            & \multicolumn{1}{c|}{50.75}    & \multicolumn{1}{c|}{53.93}     & \multicolumn{1}{c|}{51.26}     & \multicolumn{1}{c|}{52.61}     & \multicolumn{1}{c|}{37.22}     & 38.88     \\ \hline
TCBERT-base$^\spadesuit$            & \multicolumn{1}{c|}{45.82}    & \multicolumn{1}{c|}{47.06}     & \multicolumn{1}{c|}{42.91}     & \multicolumn{1}{c|}{43.87}     & \multicolumn{1}{c|}{33.28}     & 34.76     \\
TCBERT-large$^\spadesuit$           & \multicolumn{1}{c|}{50.10}    & \multicolumn{1}{c|}{50.90}     & \multicolumn{1}{c|}{53.78}     & \multicolumn{1}{c|}{53.33}     & \multicolumn{1}{c|}{37.62}     & 36.94     \\
TCBERT-1.3B$^\spadesuit$            & \multicolumn{1}{c|}{50.70}    & \multicolumn{1}{c|}{53.48}     & \multicolumn{1}{c|}{52.66}     & \multicolumn{1}{c|}{54.40}     & \multicolumn{1}{c|}{36.88}     & 38.48     \\ \hline\hline
\multirow{3}{*}{Model} & \multicolumn{6}{c}{Prompt-2}                                                                                                                                                  \\ \cline{2-7} 
                       & \multicolumn{2}{c|}{TNEWS}                                     & \multicolumn{2}{c|}{CSLDCP}                                     & \multicolumn{2}{c}{IFLYTEK}                \\ \cline{2-7} 
                       & \multicolumn{1}{c|}{referece} & \multicolumn{1}{c|}{whitening} & \multicolumn{1}{c|}{reference} & \multicolumn{1}{c|}{whitening} & \multicolumn{1}{c|}{reference} & whitening \\ \hline
Macbert-base           & \multicolumn{1}{c|}{42.29}    & \multicolumn{1}{c|}{45.22}     & \multicolumn{1}{c|}{34.23}     & \multicolumn{1}{c|}{37.48}     & \multicolumn{1}{c|}{29.62}     & 34.13    \\
Macbert-large          & \multicolumn{1}{c|}{46.22}    & \multicolumn{1}{c|}{49.60}     & \multicolumn{1}{c|}{40.11}     & \multicolumn{1}{c|}{44.26}     & \multicolumn{1}{c|}{32.36}     & 35.16     \\
Erlangshen-1.3B           & \multicolumn{1}{c|}{46.17}    & \multicolumn{1}{c|}{49.10}     & \multicolumn{1}{c|}{40.45}     & \multicolumn{1}{c|}{45.88}     & \multicolumn{1}{c|}{30.36}     & 36.88     \\ \hline
TCBERT-base$^\diamondsuit$            & \multicolumn{1}{c|}{48.31}    & \multicolumn{1}{c|}{51.34}     & \multicolumn{1}{c|}{43.42}     & \multicolumn{1}{c|}{45.27}     & \multicolumn{1}{c|}{33.10}     & 36.19     \\
TCBERT-large$^\diamondsuit$           & \multicolumn{1}{c|}{51.19}    & \multicolumn{1}{c|}{51.69}     & \multicolumn{1}{c|}{52.55}     & \multicolumn{1}{c|}{53.28}     & \multicolumn{1}{c|}{34.31}     & 37.45     \\
TCBERT-1.3B$^\diamondsuit$            & \multicolumn{1}{c|}{52.14}    & \multicolumn{1}{c|}{52.39}     & \multicolumn{1}{c|}{51.71}     & \multicolumn{1}{c|}{53.89}     & \multicolumn{1}{c|}{33.62}     & 38.14     \\ \hline
TCBERT-base$^\clubsuit$            & \multicolumn{1}{c|}{48.81}    & \multicolumn{1}{c|}{52.19}     & \multicolumn{1}{c|}{46.44}     & \multicolumn{1}{c|}{49.13}     & \multicolumn{1}{c|}{36.08}     & 37.62     \\
TCBERT-large$^\clubsuit$           & \multicolumn{1}{c|}{49.70}    & \multicolumn{1}{c|}{50.80}     & \multicolumn{1}{c|}{50.03}     & \multicolumn{1}{c|}{50.92}     & \multicolumn{1}{c|}{36.82}     & 38.99     \\
TCBERT-1.3B$^\clubsuit$            & \multicolumn{1}{c|}{50.65}    & \multicolumn{1}{c|}{53.93}     & \multicolumn{1}{c|}{50.20}     & \multicolumn{1}{c|}{52.61}     & \multicolumn{1}{c|}{36.99}     & 39.17     \\ \hline
TCBERT-base$^\spadesuit$            & \multicolumn{1}{c|}{46.72}    & \multicolumn{1}{c|}{48.86}     & \multicolumn{1}{c|}{43.19}     & \multicolumn{1}{c|}{43.53}     & \multicolumn{1}{c|}{34.08}     & 35.79     \\
TCBERT-large$^\spadesuit$           & \multicolumn{1}{c|}{50.65}    & \multicolumn{1}{c|}{51.94}     & \multicolumn{1}{c|}{53.84}     & \multicolumn{1}{c|}{53.67}     & \multicolumn{1}{c|}{37.74}     & 36.65     \\
TCBERT-1.3B$^\spadesuit$            & \multicolumn{1}{c|}{50.75}    & \multicolumn{1}{c|}{54.78}     & \multicolumn{1}{c|}{51.43}     & \multicolumn{1}{c|}{54.34}     & \multicolumn{1}{c|}{36.48}     & 38.36     \\ \hline
\end{tabular}
\caption{Testing results of sentence similarity. $\diamondsuit$, $\clubsuit$ and $\spadesuit$ denote the TCBERTs are pre-trained with prompt-based MLM, prompt-based contrastive learning and prompt-based MOCO, respectively. The difference between "reference" and "whitening" is whether the extracted representation is whitened or not.}
\label{tab:main-zero}
\end{table*}
\section{Results}
Table~\ref{tab:main-finetune} and Table~\ref{tab:main-zero} present the testing accuracy scores of the three evaluation datasets for fine-tuning and sentence similarity, respectively. We report the highest score of all fine-tuning experiments using the same seed.

As shown in Table~\ref{tab:main-finetune}, despite that different datasets may prefer different pre-training methods or prompts, we can still observe that prompt-based MLM or prompt-based MOCO for pre-training, and prompt-1 for fine-tuning are the most preferable combinations. 

Table~\ref{tab:main-zero} presents the classification results using sentence similarity. We also adopt the whitening method~\cite{su2021whitening} on the extracted representations before calculating the sentence similarity. The similarity score by whitening is denoted as "whitening" in Table~\ref{tab:main-zero} while the "reference" denotes similarity scores without whitening operations. Differing from the fine-tune results of which representation learning, i.e., contrastive learning, sentence similarity benefits from the representation learning in most cases, no matter pre-training methods or prompt versions. The whitening operation further improves the similarity scores, especially for TCBERT-1.3B.

From Table~\ref{tab:main-finetune} and Table~\ref{tab:main-zero}, we can observe noticeable performance gaps between the two classification methods. The fine-tuning scores are higher but it normally takes more time, like hours, for training. The sentence similarity only takes a few minutes to obtain the classification results once the representation is completely extracted, and the extraction is a one-time operation.

We also want to point out that for the momentum update used in MOCO pre-training, we use the two coefficients of 0.999 and 0.9999 to update the momentum encoder. We did not examine the coefficients of 0.9 and 0.99 for pre-training since the performance using these two coefficients for contrastive fine-tuning is far from comparable. To simplify, we do not decouple the size of the queue in the original MOCO design from the mini-batch size.

\section{Conclusion}
In the report, we present the pre-training methods and details for Chinese Topic Classification BERTs (TCBERTs) open-sourced at Hugging Face by Cognitive Computing and Natural Language Group, IDEA.

\bibliography{anthology,custom}
\bibliographystyle{acl_natbib}
\end{CJK}
\end{document}